\documentclass[conference]{IEEEtran}
\IEEEoverridecommandlockouts
\usepackage{cite}
\usepackage{amsmath,amssymb,amsfonts}
\usepackage{graphicx}
\usepackage{textcomp}
\usepackage{xcolor}

\usepackage{makecell}
\usepackage{color}

\usepackage{algorithm}
\usepackage{algorithmicx}

\usepackage{algpseudocode}

\def\BibTeX{{\rm B\kern-.05em{\sc i\kern-.025em b}\kern-.08em
    T\kern-.1667em\lower.7ex\hbox{E}\kern-.125emX}}
\begin{document}

\title{MaFreeI2P: A Matching-Free Image-to-Point Cloud Registration Paradigm with Active Camera Pose Retrieval
\thanks{This research is supported by the National Natural Science Foundation of
China under Grants No. U22A20102. Yu Pan is the corresponding author.}
}

\author{Gongxin Yao$^{1}$, Xinyang Li$^{2}$, Yixin Xuan$^{2}$ and Yu Pan$^{1}$\\
\IEEEauthorblockA{$^{1}$ \textit{Institute of Cyber-Systems and Control, Zhejiang University, Hangzhou, China}\\$^{2}$ \textit{Polytechnic Institute, Zhejiang University, Hangzhou, China}\\\{yaogongxin, l\_xyang, 22260380, ypan\}@zju.edu.cn
}
}

\maketitle

\begin{abstract}
Image-to-point cloud registration seeks to estimate their relative camera pose, which remains an open question due to the data modality gaps. The recent matching-based methods tend to tackle this by building 2D-3D correspondences. In this paper, we reveal the information loss inherent in these methods and propose a matching-free paradigm, named MaFreeI2P. Our key insight is to actively retrieve the camera pose in SE(3) space by contrasting the geometric features between the point cloud and the query image. To achieve this, we first sample a set of candidate camera poses and construct their cost volume using the cross-modal features. Superior to matching, cost volume can preserve more information and its feature similarity implicitly reflects the confidence level of the sampled poses. Afterwards, we employ a convolutional network to adaptively formulate a similarity assessment function, where the input cost volume is further improved by filtering and pose-based weighting. Finally, we update the camera pose based on the similarity scores, and adopt a heuristic strategy to iteratively shrink the pose sampling space for convergence. Our MaFreeI2P achieves a very competitive registration accuracy and recall on the KITTI-Odometry and Apollo-DaoxiangLake datasets.
\end{abstract}

\section{Introduction}
\label{sec:intro}
\begin{figure}[t]
    \centering
    \includegraphics[width=1.0\linewidth]{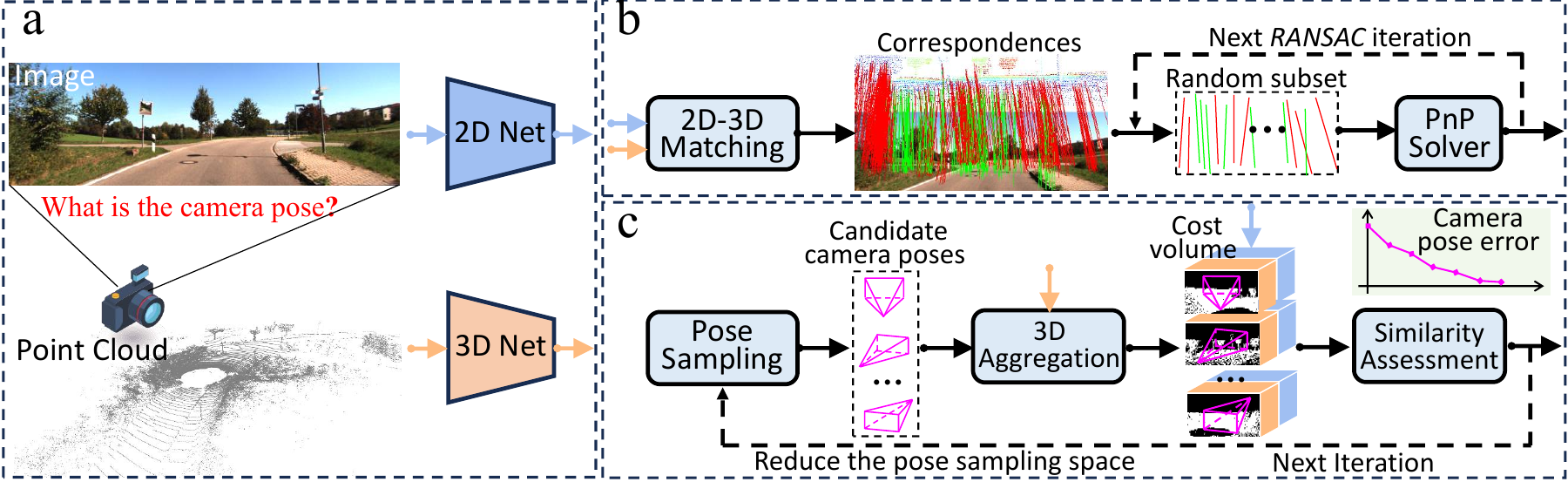}
    \caption{(a) Illustration of the image-to-point cloud registration, where the features are extracted by a pseudo-siamese neural network. (b) The mainstream matching-based pipeline. (c) Our matching-free paradigm compares the 2D features with the 3D features viewed from various camera poses, iteratively updating the output as the pose with highest similarity.}
    \label{fig1}
\end{figure}

Given a 3D scene point cloud and an RGB image capturing part of the same scene, image-to-point cloud registration aims at estimating the camera pose with respect to the coordinate frame of the 3D scene, as illustrated in Fig. \ref{fig1} (a). It has various computer vision applications, including but not limited to camera relocalization in a pre-built 3D map, 3D reconstruction, and camera-LiDAR calibration. However, RGB images capture colors and textures with 2D grid data, whereas point clouds represent 3D geometry in an unordered format. This disparity poses significant challenges to the task.

Following the matching-based approaches used for same-modal registration in images \cite{sarlin2020superglue, 2021LoFTR, Xu_2022_CVPR, xie2024deepmatcher} or point clouds \cite{lu2021hregnet, huang2021predator, qin2022geometric, wu2023sacf}, the mainstream pipeline (as illustrated in Fig. \ref{fig1} (b)) for image-to-point cloud registration is to first extract the 2D-3D correspondences with pseudo-siamese neural networks \cite{xia2021pseudo}. Subsequently, the camera pose is estimated using a RANSAC-based PnP solver \cite{fischler1981random, lepetit2009ep}. Recent algorithms such as 2D3D-MatchNet \cite{feng20192d3d}, CorrI2P \cite{ren2022corri2p} and CFI2P \cite{yao2023cfi2p} fall into this category. Although they solve the task by combining deep learning with traditional geometric vision model \cite{zheng2013revisiting}, they are still plagued by at least two issues. 1) Analogous to the concept of hard label in knowledge distillation \cite{gou2021knowledge}, a correspondence can be seen as a hard result that only retains the information of the matched point-pixel pair and discards the beneficial information between other points or pixels. 2) The RANSAC-based PnP solver eliminates correspondence outliers by iterative sampling, so the pose estimation is derived from the local perspectives on the sampled subsets.

There are also some matching-free algorithms \cite{li2021deepi2p, 9799751} that bypass building 2D-3D correspondences. DeepI2P \cite{li2021deepi2p} casts the registration process into a two-stage point classification and pose optimization framework. However, it suffers from low registration accuracy because the classification merely determines whether a 3D point is inside the image frustum, which is insufficient for precise pose fine-tuning. EFGHNet \cite{9799751} adopts a divide-and-conquer strategy to estimate rotation and translation sequentially. It projects the point cloud onto a 360° panoramic depth map and then seeks out the shared geometric features between this map and the image. Thus, EFGHNet is confined to scenarios with small translation, since significant movement would destroy the shared geometric features. So far, these matching-free methods significantly lagged behind their matching-based counterparts.

\begin{figure}[t]
    \centering
    \includegraphics[width=1.0\linewidth]{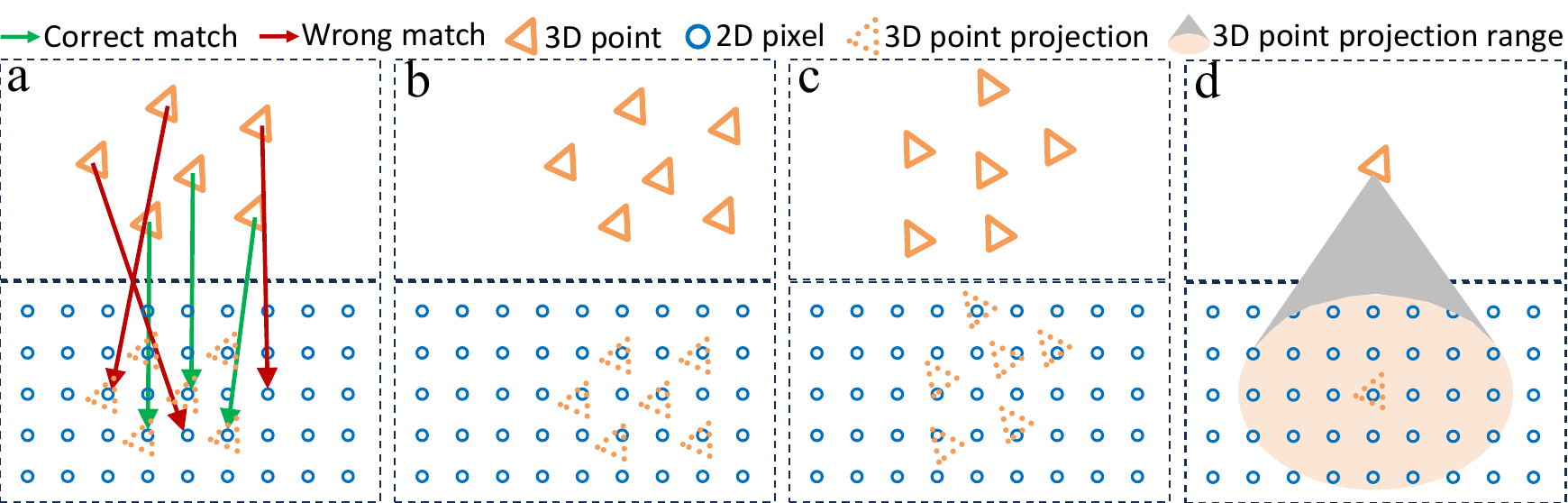}
    \caption{The matching-based pipelines only retain the information of 2D-3D matches even if some of them are wrong as shown in (a). Cost volume builds virtual correspondences between 3D points and their projected 2D pixels. Each cost volume unit projects 3D points using a distinct camera pose, as shown in (b) and (c). Consequently, the information between each 3D point and all 2D pixels within its total projection range can be preserved in the whole cost volume, as shown in (d). Besides, projection partially maintains the spatial proximity of 3D points, whereas the wrong matches destroy it.
    }
    \label{fig2}
\end{figure}

In this paper, we introduce MaFreeI2P, a novel \textbf{Ma}tching-\textbf{Free} \textbf{I}mage-\textbf{to}-\textbf{P}oint cloud registration paradigm that overcomes the above limitations. We still adopt a pseudo-siamese network as the backbone to learn 2D and 3D features. Inspired by the success of cost volume for depth estimation \cite{tian2023dps} and stucture-from-motion \cite{wei2023deepsfm}, we fuse the cross-modal features to construct a pose-based cost volume, rather than establishing 2D-3D correspondences that lead to substantial information loss. Specifically, we first uniformly sample a set of discrete candidate camera poses around the current or initial pose, and use them to project the point cloud into various 2D views. For each sampled camera pose, the 3D features are aggregated according to its corresponding 2D view, and then concatenated with 2D features to form an individual cost volume unit. All the units are fed into a convolutional network to calculate the similarity scores, and we update the current camera pose as the one with highest score. The above process of pose sampling and similarity assessment is repeated iteratively, as shown in Fig. \ref{fig1} (c). Such intuitive pipeline has at least three advantages. 1) It has excellent interpretability, reflecting the instinctive process of humans in adjusting and observing to align two objects. 2) By building cost volume, it can retain and exploit more information than the matching-based pipeline as illustrated in Fig. \ref{fig2}. 3) Since no RANSAC strategy is applied to the points and pixels, they are all taken into global consideration within each cost volume unit.

Beyond the basic paradigm, to guarantee convergence and achieve a global optimum, we emulate classic gradient descent algorithms for sampling camera poses. It begins with an initially large sampling space and step size, which are then reduced iteratively. To improve the feature quality of the basic cost volume, we opt to identify the 3D points outside the image frustum and zero out the corresponding features. Furthermore, to focus on the 2D and 3D features that are more critical within the cost volume, we assign pose-based weights to different regions, thereby differentiating their importance in the computation of similarity scores.

In summary, our main contributions include: 1) We propose a novel matching-free image-to-point cloud registration paradigm, retrieving the camera pose in SE(3) space via active sampling and similarity assessment. 2) We build camera pose-based cost volume with the cross-modal features, preserving richer information than 2D-3D matching, where each cost unit implicitly reflects the confidence level of a sampled pose. 3) We design a convolutional network to adaptively formulate the similarity assessment function, where the input cost volume is further improved by filtering and pose-based weighting. 4) We achieve state-of-the-art performance in registration accuracy and recall on the KITTI-Odometry \cite{geiger2013vision} and Apollo-DaoxiangLake \cite{zhou2020da4ad} datasets.

\section{METHODOLOGY}
\subsection{Problem Formulation}
\begin{figure*}[t]
    \centering
    \includegraphics[width=1\linewidth]{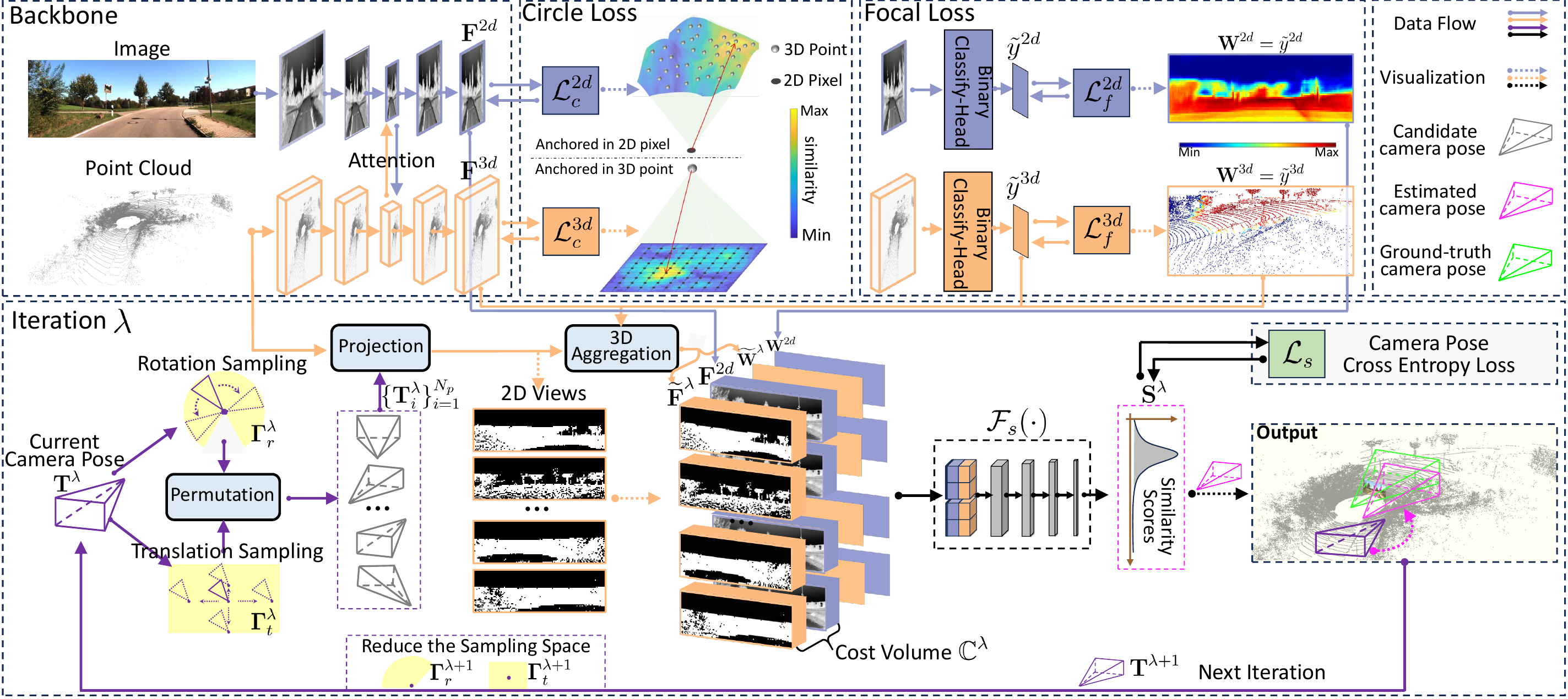}
    \caption{The architecture of MaFreeI2P. \textbf{Top:} The backbone and classification heads are supervised by the circle loss in Eq. \ref{eq2} and the focal loss in Eq. \ref{eq5}, respectively. \textbf{Bottom:} The iterative branch in Section \ref{sec:sampling}$\sim$\ref{sec:similarity}. $\mathcal{F}_{s}(\cdot)$ is a light-weight network supervised by the cross-entropy loss in Eq. \ref{eq8}.
    }
    \label{fig3}
\end{figure*}

Given a scene point cloud $\textbf{P} \in \mathbb{R}^{N \times 3}$ with $N$ 3D points and a query RGB image $ \textbf{I} \in \mathbb{R}^{H' \times W' \times 3} $ with a resolution of $H' \! \times \! W'$, our goal is to estimate the pose of the camera that captured $\textbf{I}$ within $\textbf{P}$ as shown in Fig. \ref{fig1} (a). Mathematically, the camera pose is a rigid transformation $\overline{\textbf{T}} = [\bar{\textbf{R}}|\bar{\textbf{t}}] $ with a rotation $\bar{\textbf{R}} \in SO(3)$ and a translation $\bar{\textbf{t}} \in \mathbb{R}^{3}$, which aligns the projection of 3D points with the correlated 2D pixels as:
\begin{align}
    \label{eq1}
	\overline{\textbf{T}} &= \mathop{\mathrm{arg\,min}}\limits_{\textbf{T} = [\textbf{R}|\textbf{t}]} \sum\nolimits_{(\textbf{u}_{i},\textbf{p}_{j}) \in M^{*}} {\Vert \textbf{u}_{i} - \mathcal{K}(\textbf{T}\textbf{p}_{j})\Vert}_{2}\,,
\end{align}
where $M^{*}$ is the 2D-3D correspondence set, $\textbf{u}_{i}$ and $\textbf{p}_{j}$ are a pair of matched pixel and point, $\textbf{T}\textbf{p}_{j}\!$ transforms the 3D point $\textbf{p}_j$ (i.e., $\textbf{R}\textbf{p}_{j} + \textbf{t}$), and $\mathcal{K}(\cdot)$ is the camera projection function. We argue that it is difficult to establish precise correspondence set via matching, which also leads to a significant information loss. In the next sections, we will describe the details of our matching-free solution illustrated in Fig. \ref{fig3}.

\subsection{Point and Pixel Geometric Embedding}
\label{sec:feat}
To construct cross-modal features that bridges the modality gap between images and point clouds, we adopt a pseudo-siamese network with a multi-scale architecture as the backbone to map both $\textbf{I}$ and $\textbf{P}$ into a unified embedding space. Besides, attention mechanism is performed on the down-sampled features to exchange cross-modal information. The final 2D and 3D features are denoted as $ \textbf{F}^{2d} \in \mathbb{R}^{H \times W \times f} $ and $ \textbf{F}^{3d} \in \mathbb{R}^{N \times f}$, where $f$ represents the feature dimension.


We construct the circle loss \cite{sun2020circle} between 3D points and 2D pixels to supervise the backbone training. For this purpose, we start by sampling the points that can be projected within the image frustum. The sampled 3D points form a subset $\textbf{P}^{s} \in \mathbb{R}^{N_{s} \times 3}$ and the corresponding 2D pixel subset is denoted as $\textbf{U}^{s} = \{\textbf{u}_i \big| \textbf{u}_i =  \mathcal{K}(\overline{\textbf{T}}\textbf{p}_{i}), \textbf{p}_{i} \in \textbf{P}^{s}\}$. For any pixel $\textbf{u}_i \in \textbf{U}^{s}$ , the set of positive 3D points is defined as $\mathcal{E}_{p}^{i} = \{\textbf{p}_{j} \,\big|\, \textbf{p}_{j} \in \textbf{P}^{s}, {\Vert \textbf{u}_{i} - \mathcal{K}(\overline{\textbf{T}}\textbf{p}_{j})\Vert}_{2} \leq r  \}$, where $r$ is a radius. Similarly, the set of negative 3D points is defined as $\mathcal{E}_{n}^{i} = \{\textbf{p}_{j} \,\big|\, \textbf{p}_{j} \in \textbf{P}^{s}, {\Vert \textbf{u}_{i} - \mathcal{K}(\overline{\textbf{T}}\textbf{p}_{j})\Vert}_{2} > r \}$. The circle loss anchored in 2D pixels is then computed as: 
\begin{align}
    \label{eq2}
    \mathcal{L}^{2d}_{c} = \frac{1}{N_s} \sum_{i=1}^{N_s} \log \left[ 1 \! + \! \sum_{j\in\mathcal{E}_{p}^{i}} e^{\theta_p^j(d_i^j - \Delta_p)} \cdot \! \sum_{k\in\mathcal{E}_{n}^{i}} e^{\theta_n^k(\Delta_n - d_i^k)} \right],
\end{align}
where $d_i^j \! = \! {\Vert \textbf{F}^{2d}_{\textbf{u}_{i}} -  \textbf{F}^{3d}_{\textbf{p}_{j}} \Vert}_{2}$ denotes the feature distance between the 2D pixel $\textbf{u}_{i}$ and 3D point $\textbf{p}_{j}$, and $\Delta_p$, $\Delta_n$ are the positive and negative margins, respectively. The adaptive weights $\theta_p^j = \gamma(d_i^j \! - \! \Delta_p)$ and $\theta_n^k = \gamma(\Delta_n \! - \! d_i^k)$ are computed individually for each positive and negative samples. In experiment, we set $\gamma:=10$, $\Delta_p:=0.1$ and $\Delta_n:=1.4$ empirically. The circle loss anchored in 3D points $\mathcal{L}^{3d}_{c}$ is computed in the same way, and the total circle loss is defined as $\mathcal{L}_{c} \! = \! \mathcal{L}^{2d}_{c} \! + \! \mathcal{L}^{3d}_{c}$. This supervision optimizes the 2D and 3D features from two directions, as illustrated in the circle loss part of Fig. \ref{fig3}.

\subsection{Camera Pose Sampling}
\label{sec:sampling}
We sample a set of camera poses to construct the pose-based cost volume. At each iteration $\lambda$, denote the current camera pose as $\textbf{T}^{\lambda} = [\textbf{R}^{\lambda}|\textbf{t}^{\lambda}]$, the current translation sampling space as $\boldsymbol{\Gamma}^{\lambda}_t$, and the current rotation sampling space as $\boldsymbol{\Gamma}^{\lambda}_r$. For translation, we sample $\{ \delta \textit{\textbf{t}}_i \}_{i=1}^{N_t}$ uniformly in $\boldsymbol{\Gamma}^{\lambda}_t$ and add each $\delta \textit{\textbf{t}}_i$ to the current $\textbf{t}^{\lambda}$, where $\delta \textit{\textbf{t}}_i$ is a 3D vector. For rotation, we uniformly sample a batch of Euler angles according to $\boldsymbol{\Gamma}^{\lambda}_r$ and convert them into rotation matrices $\{ \delta \textit{\textbf{R}}_i \}_{i=1}^{N_r}$, where each $\delta \textit{\textbf{R}}_i$ is multiplied by the current $\textbf{R}^{\lambda}$. As depicted in the bottom-left of Fig.\ref{fig3}, we combine the rotation and translation to generate a group of candidate poses $\{ \textbf{T}^{\lambda}_i \}_{i=1}^{N_p}$, where $N_p = N_r \cdot N_t$.

\subsection{Pose-based Cost Volume}
\label{sec:cost}
We construct cost volume with the cross-modal features in Section. \ref{sec:feat} and the sampled camera poses in Section. \ref{sec:sampling}. For the $i$-th candidate camera pose $\textbf{T}^{\lambda}_i$, we first allocate an empty placeholder $\widetilde{\textbf{F}}^{\lambda i} \in \mathbb{R}^{H \times W \times f}$ with the same size of the image features $\textbf{F}^{2d}$. Then we project the raw point cloud $\textbf{P}$ with $\textbf{T}^{\lambda}_i$, aggregate the 3D features for the projected points, and compute mean vectors to fill $\widetilde{\textbf{F}}^{\lambda i}$ as:
\begin{small}
\begin{align}
    \label{eq3}
    \widetilde{\textbf{F}}^{\lambda i}_{\tilde{\textbf{u}}} \! = \! \sum\limits_{\textbf{p}_{j} \in G^{\lambda i}_{\tilde{\textbf{u}}}} \frac{\textbf{F}^{3d}_{\textbf{p}_{j}}}{|G^{\lambda i}_{\tilde{\textbf{u}}}|}\,\,\,\mathrm{s.t.}\,\,\, G^{\lambda i}_{\tilde{\textbf{u}}} \! = \! \left\{ \textbf{p}_{j} \in \textbf{P} \,\big|\, \tilde{\textbf{u}} =  \mathcal{K}(\textbf{T}^{\lambda}_i\textbf{p}_{j})\right\},
\end{align}
\end{small}where $\tilde{\textbf{u}}$ represents a 2D position to index the pixel in $\widetilde{\textbf{F}}^{\lambda i}$, and $|\cdot|$ denotes set cardinality. We initialize $\widetilde{\textbf{F}}^{\lambda i}_{\tilde{\textbf{u}}}$ with zero if there are no point in $\textbf{P}$ that is projected at the pixel $\tilde{\textbf{u}}$.

We build a 4D cost volume $\mathbb{C}^{\lambda} \in \mathbb{R}^{N_p \times H \times W \times 2f}$ with all candidate camera poses, where a single cost volume unit is built by concatenating the image features $\textbf{F}^{2d}$ with $\widetilde{\textbf{F}}^{\lambda i}$ as:
\begin{align}
    \label{eq4}
    \mathbb{C}^{\lambda}_i \! = \! \mathrm{Cat}(\,\textbf{F}^{2d},\, \widetilde{\textbf{F}}^{\lambda i}\,).
\end{align}
In each cost volume unit, the aggregated 3D feature at each pixel will be compared with the corresponding 2D feature. It can be assumed that there is a virtual correspondence between $\textbf{p}_{j}$ and $\tilde{\textbf{u}}$ according to Eq. \ref{eq3}. Thus, more information can be preserved in the whole cost volume, as illustrated in Fig. \ref{fig2}.

\subsection{Similarity Assessment}
\label{sec:similarity}
We will compute a similarity score for each cost volume unit. However, two critical aspects have been overlooked so far. 1) The circle loss in Eq. \ref{eq2} only supervises the 3D points inside the image frustum, so the low-quality features of points outside the image frustum will contaminate the cost volume. 2) The basic cost volume unit in Eq. \ref{eq3} lacks a mechanism to discriminate the contribution of different pixels when calculating the overall similarity score. To mitigate these issues, we design the similarity assessment strategy as follows:

\vspace{1mm}
\textbf{Zeroing Out the Inferior Features.} We first identify the 3D points inside the image frustum and the 2D pixels where at least one 3D point is projected. Thus, two binary classification heads are added to the end of the backbone, predicting the confidence scores $\tilde{y}^{2d} \! \in \! \mathbb{R}^{H \times W \times 1}$ and $\tilde{y}^{3d} \! \in \! \mathbb{R}^{N\times1}$ of image and point cloud respectively. Considering the imbalanced number of the points inside and outside the image frustum, we construct a focal loss \cite{lin2017focal} to supervise the 3D head training:
\begin{align}
    \label{eq5}
    \mathcal{L}_{f}^{3d} = \mathrm{FocalLoss}(\tilde{y}^{3d}, y^{3d}),
\end{align}
where $y^{3d}_i$ is 1 if $\textbf{p}_i$ is inside the image frustum and otherwise 0. $\mathcal{L}_{f}^{2d}$ is computed in the same way and the total focal loss is defined as $\mathcal{L}_{f} \! = \! \mathcal{L}_{f}^{2d} \! + \! \mathcal{L}_{f}^{3d}$. In inference, we only zero out the 3D feature $\textbf{F}^{3d}_{\textbf{p}_{j}}$ if $\tilde{y}^{3d}_{\textbf{p}_{j}} < 0.5$\,. We do not zero out any 2D feature as our purpose is to push the inferior 3D features away from the 2D features. Zeroing out one of them is enough. 


\vspace{1mm}
\textbf{Pose-based Weighting.} The above confidence scores also reflect the significance of 3D points and 2D pixels. A higher score means that the corresponding feature should contribute more to similarity assessment. Therefore, we reuse $\tilde{y}^{2d}$ and $\tilde{y}^{3d}$ as the weights $\textbf{W}^{2d} \in \mathbb{R}^{H \times W \times 1}$ and $\textbf{W}^{3d} \in \mathbb{R}^{N \times 1}$ for the image and point cloud respectively. For the $i$-th sampled camera pose $\textbf{T}^{\lambda}_i$ at the $\lambda$-th iteration, we also allocate an empty placeholder $\widetilde{\textbf{W}}^{\lambda i} \in \mathbb{R}^{H \times W \times 1}$. Similar to the cost volume, we aggregate 3D weights to fill the placeholder as:
\begin{small}
\begin{align}
    \label{eq6}
    \widetilde{\textbf{W}}^{\lambda i}_{\tilde{\textbf{u}}} \! = \! \sum\limits_{\textbf{p}_{j} \in G^{\lambda i}_{\tilde{\textbf{u}}}} \! \textbf{W}^{3d}_{\textbf{p}_{j}} \,\,\,\,\mathrm{s.t.}\,\,\,\, G^{\lambda i}_{\tilde{\textbf{u}}} \! = \! \left\{ \textbf{p}_{j} \in \textbf{P} \,\big|\, \tilde{\textbf{u}} =  \mathcal{K}(\textbf{T}^{\lambda}_i\textbf{p}_{j})\right\}.
\end{align}
\end{small}In the end, $\textbf{W}^{2d}$ and $\widetilde{\textbf{W}}^{\lambda i}$ are utilized to weight $\textbf{F}^{2d}$ and $\widetilde{\textbf{F}}^{\lambda i}$ respectively in each cost volume unit $\mathbb{C}^{\lambda}_i$.


\vspace{1mm}
\textbf{Similarity Score Calculation.} For each cost volume unit $\mathbb{C}^{\lambda}_i$, an intuitive similarity score is the sum of the $H\times W$ pixel-wise distances between the 2D features and the aggregated 3D features. However, this score overlooks the noise and the spatial correlations among adjacent pixels. To address this, we employ a light-weight network with only 10 2D conv layers to formulate the similarity estimation function $\mathcal{F}_{s}(\cdot)$. As for weighting, it is inappropriate to directly scale $\textbf{F}^{2d}$ and $\widetilde{\textbf{F}}^{\lambda i}$ with the weights $\textbf{W}^{2d}$ and $\widetilde{\textbf{W}}^{\lambda i}$. Instead, we concatenate the weights with the cost volume unit as the input, allowing $\mathcal{F}_{s}(\cdot)$ to evolve into a function that can utilize the weights effectively. So the similarity score $\textbf{S}^{\lambda}_i$ is calculated as:
\begin{align}
    \label{eq7}
    \textbf{S}^{\lambda}_i = \mathcal{F}_{s}(\mathbb{C}^{\lambda}_i, \,\textbf{W}^{2d}, \widetilde{\textbf{W}}^{\lambda i}).
\end{align}
All the cost volume units $\mathbb{C}^{\lambda}$ can be processed in parallel to get a batch of scores $\textbf{S}^{\lambda} \in \mathbb{R}^{N_p}$. To train $\mathcal{F}_{s}(\cdot)$, a simple cross entropy loss is adopted to supervise $\textbf{S}^{\lambda}$ as:
\begin{align}
    \label{eq8}
    \mathcal{L}_{s} = \mathrm{CrossEntropy}(\textbf{S}^{\lambda}, \Phi^{\lambda}),
\end{align}
where $\Phi^{\lambda} \in \mathbb{R}^{N_p}$ are the binary indicators, in which $\Phi^{\lambda}_{i}$ is 1 if the sampled camera pose $\textbf{T}^{\lambda}_i$ is closest to the ground truth pose $\overline{\textbf{T}}$ otherwise 0.


\subsection{Inference Iteration}
\label{sec:inference}
As illustrated in Fig.\ref{fig3}, we iteratively perform the operations in Section \ref{sec:sampling}$\sim$\ref{sec:similarity}. At each iteration, we update the current camera pose as the one with highest similarity score, which serves as the new center to sample candidate camera poses around it in the next iteration. To guarantee convergence and achieve a global optimum, we gradually reduce the pose sampling space (i.e., $\boldsymbol{\Gamma}^{\lambda}_t$ and $\boldsymbol{\Gamma}^{\lambda}_r$) as well as the sampling step size. Since the similarity assessment for different cost volume units are independent of each other as in Eq. \ref{eq7}, we can handle the process in segments when there are too many candidate camera poses or memory is limited.

\section{Experiments}
\subsection{Experimental Settings}
\textbf{Dataset.} We conducted experiments on the KITTI-Odometry \cite{geiger2013vision} and Apollo-DaoxiangLake \cite{zhou2020da4ad} datasets. In KITTI-Odometry, the data sequences 0-8 and 9-10 are used for training and testing respectively. Apollo-DaoxiangLake captures data sequences of the area in various weather and seasons. We follow its official training and testing split, but remove data in foggy weather since the 3D data is too poor in this case. For both datasets, each point cloud is first processed by voxel downsampling with a voxel size of 0.1m, and then we randomly sample 40960 points as input. The input image size is set to $160\times512$. For a fair comparison, we follow the same protocol in the previous works \cite{ren2022corri2p, yao2023cfi2p}
to get image-point cloud pairs. Specifically, to ensure that the image contains a part of the same scene in the point cloud, we select the image and point cloud with same frame ID. To construct the registration problem, we randomly transform the point cloud with a rotation around the up-axis within a range of 360° and a translation on the ground within a range of 10 meters.

\vspace{1mm}
\textbf{Evaluation Metric.} We evaluate the registration performance with three metrics. 1) Relative translation error (RTE). 2) Relative rotation error (RRE). 3) Registration recell (RR). RR represents the fraction of successful registrations among the test samples. A registration is considered as successful if RTE is smaller than $\tau_{t}$ and RRE is smaller than $\tau_{r}$.

\vspace{1mm}
\textbf{Implementation Details.} We adopt the PyTorch 1.12 and an RTX-3090 GPU for training. Indeed, the iterative branch of MaFreeI2P is agnostic to the embedding backbone. Better embeddings and confidence scores will improve the quality of pose-based cost volume. Thus, we first train the backbone with the circle loss and focal loss for 32 epochs on the two datasets. After that, the similarity estimation network $\mathcal{F}_{s}(\cdot)$ can achieve rapid convergence with just 2 epochs of training. We utilize the ADAM optimizer with an initial learning rate of 0.001 and decrease the learning rate by a factor of 0.8 every 3 epochs. During the training of $\mathcal{F}_{s}(\cdot)$, we employed various camera pose sampling ranges and step sizes to encompass the pose convergence states of the inference iterations. Considering the registration problem setting and the GPU memory, we sample 9 angles and $9\times9$ positions around the current camera pose at each iteration, thus generating 729 candidate poses.

\begin{table}[t]
\centering
\footnotesize
\caption{Quantitative comparison on KITTI-Odometry. The mark \textcolor{cyan}{*} means that the method is matching-based.}
\label{tab1}
\begin{tabular}{c|c|c|c}
\Xhline{1pt}
 Methods     & RR(\%)    & RTE(m)         & RRE(°)         \\ \Xhline{1pt}
 DeepI2P(3D)\cite{li2021deepi2p} & 38.34 & 1.206$\pm$0.748 & 6.083$\pm$2.294 \\ 
 DeepI2P(2D)\cite{li2021deepi2p} & 74.50 & 1.417$\pm$0.897 & 3.877$\pm$2.170 \\ 
 EFGHNet \cite{9799751}          & 23.92 & 3.191$\pm$1.132 & 4.953$\pm$2.501 \\ \hline
 GridCls \cite{li2021deepi2p} \textcolor{cyan}{*}    & 80.18 & 1.137$\pm$0.696 & 5.883$\pm$2.013 \\ 
 CorrI2P \cite{ren2022corri2p} \textcolor{cyan}{*}     & 92.19 & 0.902$\pm$0.270 & 2.720$\pm$2.170 \\ 
 CFI2P \cite{yao2023cfi2p} \textcolor{cyan}{*}      & 99.37 & 0.594$\pm$0.476 & 1.395$\pm$1.076 \\ \hline 
 MaFreeI2P(Ours)   & \textcolor{red}{\textbf{99.82}} & \textcolor{red}{\textbf{0.311$\pm$0.211}} & \textcolor{red}{\textbf{1.038$\pm$1.092}} \\ \Xhline{1pt}
\end{tabular}
\end{table}

\begin{table}[t]
\centering
\footnotesize
\caption{Quantitative comparison on Apollo-DaoxiangLake. The mark \textcolor{cyan}{*} means that the method is matching-based.}
\label{tab2}
\begin{tabular}{c|c|c|c}
\Xhline{1pt}
 Methods     & RR(\%)    & RTE(m)         & RRE(°)         \\ \Xhline{1pt}
 DeepI2P(3D)\cite{li2021deepi2p} & 31.83 & 2.322$\pm$0.835 & 6.747$\pm$2.346 \\ 
 DeepI2P(2D)\cite{li2021deepi2p} & 73.83 & 2.275$\pm$1.050 & 4.017$\pm$1.994 \\ 
 EFGHNet \cite{9799751}          & 35.25 & 3.318$\pm$1.593 & 5.142$\pm$3.039 \\ \hline 
 GridCls \cite{li2021deepi2p} \textcolor{cyan}{*}     & 79.51 & 2.072$\pm$1.437 & 5.438$\pm$1.834 \\ 
 CorrI2P \cite{ren2022corri2p} \textcolor{cyan}{*}    & 89.46 & 1.259$\pm$0.903 & 1.874$\pm$1.673 \\ 
 CFI2P \cite{yao2023cfi2p} \textcolor{cyan}{*}      & \textcolor{red}{\textbf{93.23}} & 1.742$\pm$1.097 & 1.530$\pm$0.991 \\ \hline 
 MaFreeI2P(Ours)   & 90.55 & \textcolor{red}{\textbf{0.471$\pm$0.527}} & \textcolor{red}{\textbf{0.576$\pm$1.059}} \\ \Xhline{1pt}
\end{tabular}
\end{table}

\begin{figure*}[t]
    \centering
    \includegraphics[width=1.0\linewidth]{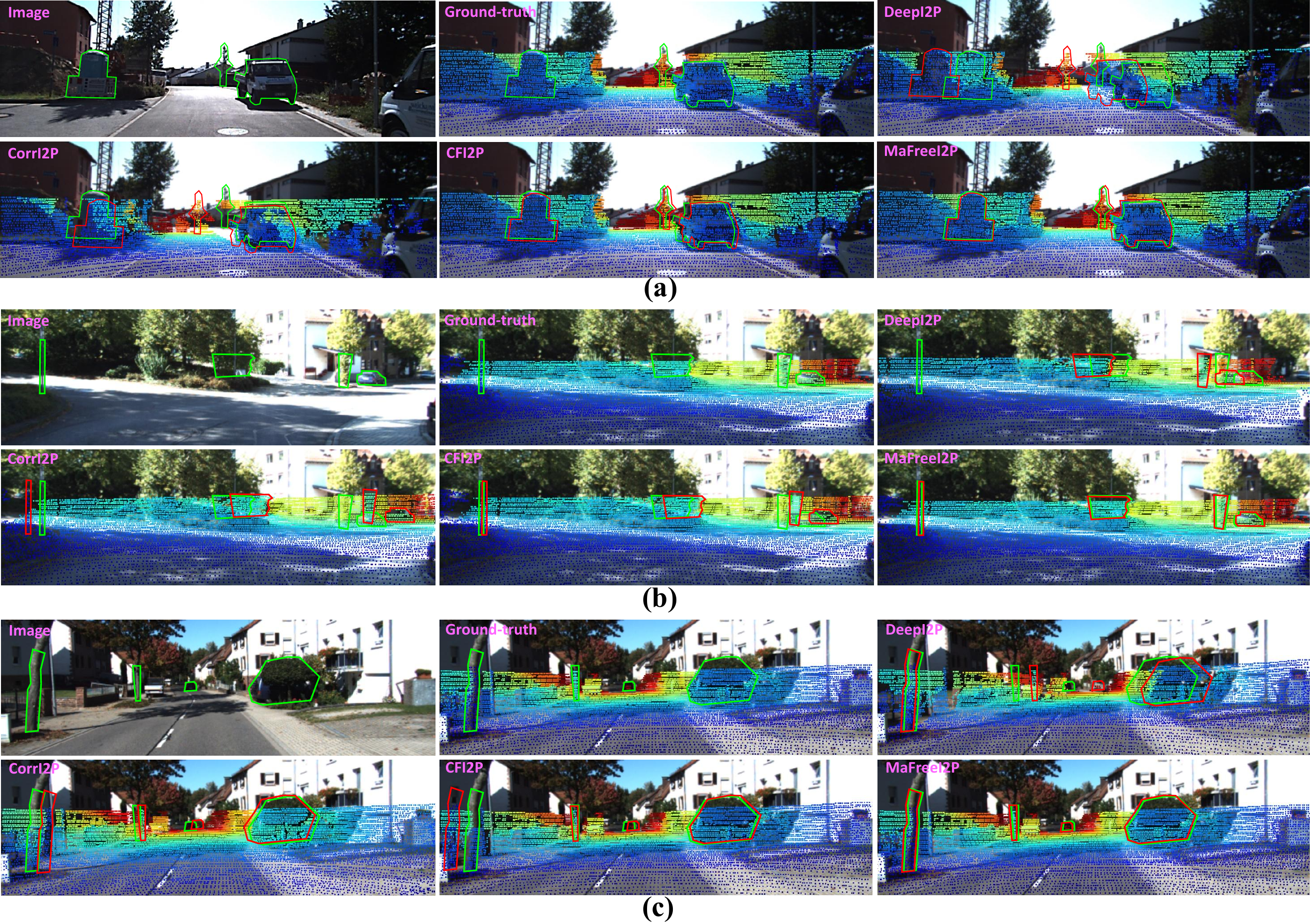}
    \vspace{-5mm}
    \caption{Three visualization examples of image-to-point cloud registration. Green outlines highlight the position of objects on the 2D image. Red outlines highlight the projection of 3D objects in the point cloud, which is transformed by the estimated camera poses. The degree of alignment between the green and red outlines reflects the registration accuracy.
    }
    \label{fig4}
\end{figure*}

\begin{figure}[t]
    \centering
    \includegraphics[width=1.0\linewidth]{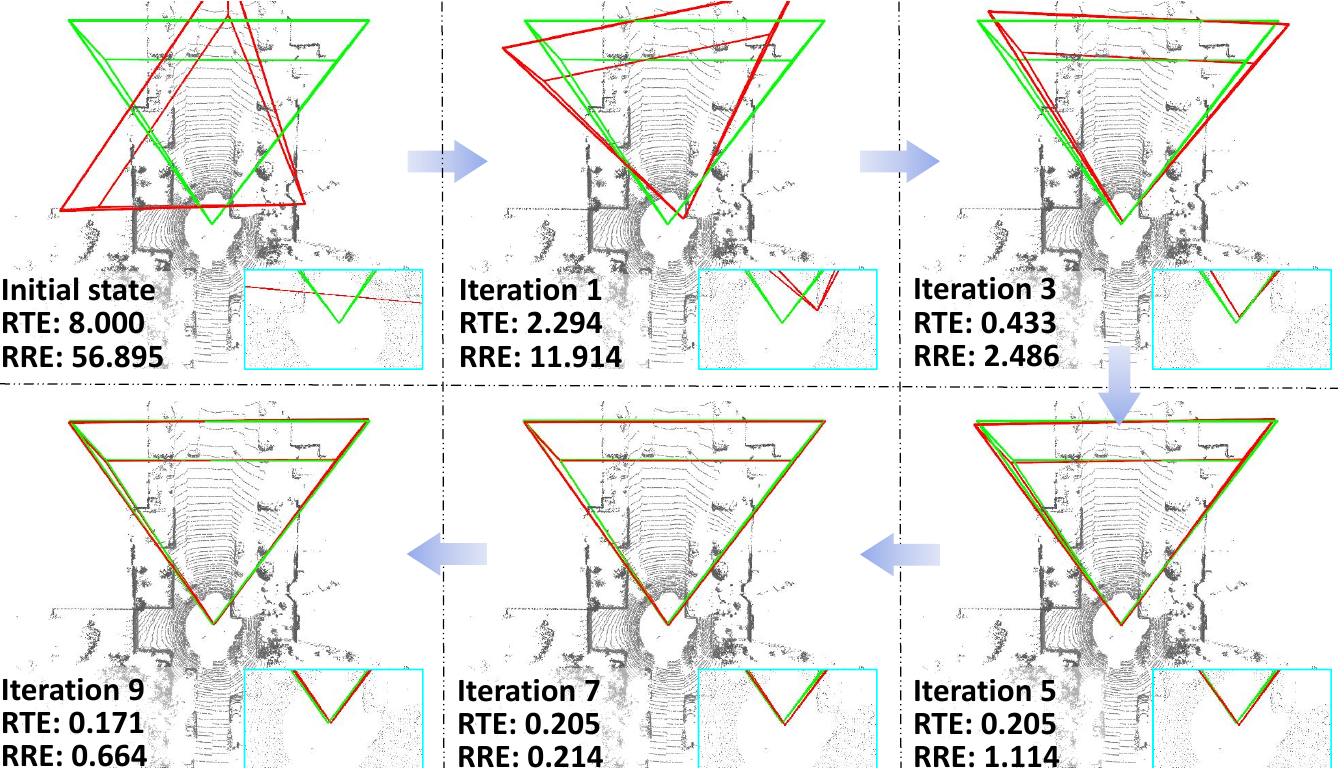}
    \vspace{-5mm}
    \caption{Visualization of MaFreeI2P iterations. The ground-truth (green) and estimated (red) camera poses are visualized in the bird's-eye view of the scene point cloud (gray).
    }
    \label{fig5}
\end{figure}

\subsection{Experimental Results}

\textbf{Comparison.} We report the quantitative results in Table \ref{tab1} and \ref{tab2}. Following the setting of previous works \cite{ren2022corri2p, yao2023cfi2p}, the thresholds $\tau_{r}$ and $\tau_{t}$ of RR are set to 10° and 5m. RTE and RRE are averaged over the test samples that satisfy the above constraints. On KITTI-Odometry, MaFreeI2P has achieved state-of-the-art results on all three metrics. The impressive RR result, reaching up to 99.82\%, demonstrates its robustness. On Apollo-DaoxiangLake, the performance of all methods degrades due to weather and season changing. Although our RR is lower than that of CFI2P, we achieve the lowest RRE and RTE, outperforming others by a large margin. Furthermore, it is also worth noting that our method is the only matching-free approach that rivals and even surpasses the performance of advanced matching-based methods. We also provide three more intuitive visual examples in Fig. \ref{fig4}. 

\vspace{1mm}
\textbf{How Does MaFreeI2P Work?} As discussed in Section \ref{sec:inference}, MaFreeI2P employs an iterative process akin to the classic gradient descent algorithm, progressively refining the initial camera pose towards the optimum. In experiments, the rotation sampling range gradually shrinks from 360° to less than 1° and the translation sampling range shrinks from 10m to less than 0.5m. We display the iterative convergence process of the camera pose in Fig \ref{fig5}.

\begin{table}[t]
\centering
\footnotesize
\caption{Ablation Studies on the KITTI-Odometry dataset. ZOIF represents the operation to zero out the inferior features. ZOIF, $\textbf{W}^{2d}$ and $\textbf{W}^{3d}$ are defined in Section \ref{sec:similarity}.}
\label{tab3}
\begin{tabular}{l|c|c|c}
\Xhline{1pt}
 \quad\quad\quad\quad\quad Methods     & RR(\%)    & RTE(m)         & RRE(°)          \\ \Xhline{1pt}
 Baseline     & 92.21 & 0.333 & 1.196 \\ 
 Baseline \! + \! ZOIF     & 99.26 & 0.390 & 1.280 \\ 
 Baseline \! + \! ZOIF \! + \! $\textbf{W}^{2d}$    & 99.80 & 0.371 & 1.106 \\ 
 Baseline \! + \! ZOIF \! + \! $\textbf{W}^{2d}$ \! + \!  $\textbf{W}^{3d}$   & 99.82 & 0.311 & 1.038 \\ \Xhline{1pt}
 
 Full pipeline with 1 iterations & 45.11 & 1.079 & 4.899 \\ 
 Full pipeline with 3 iterations & 98.99 & 0.726 & 3.817 \\ 
 Full pipeline with 5 iterations & 99.78 & 0.353 & 1.390 \\ 
 Full pipeline with 7 iterations & 99.80 & 0.314 & 1.063           \\ 
 Full pipeline with 9 iterations & 99.82 & 0.311 & 1.038 \\ \Xhline{1pt}
\end{tabular}
\end{table}

\vspace{1mm}
\textbf{Ablation Studies.} We mainly investigate the iterative branch of MaFreeI2P and report the quantitative results in Table \ref{tab3}. Firstly, individual components of the similarity assessment module are sequentially removed from the full pipeline, while maintaining the total number of iterations at 9. By this way, we find that each component provides a performance increment for MaFreeI2P. Secondly, we evaluate the full pipeline with different numbers of iterations, while the initial sampling space of pose camera remains consistent. As the iteration number increases, the final pose sampling space and steps are further refined, resulting in better performance.


\section{Conclusion}
In this paper, we propose a novel matching-free paradigm for image-to-point cloud registration named MaFreeI2P. Instead of building hard 2D-3D correspondences that lead to substantial information loss, we sample a batch of camera poses and then build a cost volume to select the best candidate with highest similarity score. MaFreeI2P is an iterative method that repeats pose sampling and similarity assessment. To ensure convergence and achieve global optimum, we emulate classic gradient descent algorithms to refine the sampled camera poses progressively. Furthermore, we have also designed a similarity assessing strategy that considers both noise and spatial correlations. The experiments on two public datasets and the impressive visualization demonstrate the effectiveness and superiority of MaFreeI2P.

\bibliographystyle{IEEEbib}
\bibliography{icme2023template}

\end{document}